\documentclass[10pt,twocolumn]{article}

\usepackage[utf8]{inputenc}
\usepackage[T1]{fontenc}
\usepackage[margin=0.75in]{geometry}
\usepackage[expansion=false]{microtype}
\usepackage{graphicx}
\graphicspath{{paper_assets/}}
\usepackage{booktabs}
\usepackage{amsmath, amssymb, amsthm}
\usepackage{mathtools}
\usepackage{natbib}
\usepackage{hyperref}
\usepackage{url}
\usepackage{xcolor}
\usepackage{xspace}
\usepackage{caption}
\usepackage{subcaption}
\usepackage{algorithm}
\usepackage{algpseudocode}
\usepackage{multirow}
\usepackage{placeins}
\usepackage{cuted}
\usepackage[font=small,labelfont=bf]{caption}
\usepackage{listings}

\lstdefinestyle{pystyle}{
  language=Python,
  basicstyle=\ttfamily\scriptsize,
  keywordstyle=\color{blue!60!black}\bfseries,
  commentstyle=\color{green!40!black}\itshape,
  stringstyle=\color{orange!60!black},
  numbers=left,
  numberstyle=\tiny\color{gray!70},
  numbersep=6pt,
  xleftmargin=1.5em,
  showstringspaces=false,
  breaklines=true,
  breakatwhitespace=false,
  frame=single,
  rulecolor=\color{gray!50},
  tabsize=4,
  captionpos=b,
  columns=flexible,
}
\hypersetup{
  colorlinks=true,
  linkcolor=blue!50!black,
  citecolor=blue!50!black,
  urlcolor=blue!50!black,
}

\newcommand{\dshift}{$D_{\text{shift}}^{n{=}50}$\xspace}

\newcommand{\embedstd}{$\sigma_{\text{embed}}$\xspace}
\newcommand{\epmjepa}{EPM-JEPA\xspace}
\newcommand{\eijepa}{EI-JEPA\xspace}
\newcommand{\vjepa}{Vanilla JEPA\xspace}
\newcommand{\lora}{LoRA\xspace}

\title{\textbf{EPM-JEPA: Operator-Side Experience Modulation in JEPA-Family World Models}}

\author{\textbf{Vedant Pandya} \\
        \textbf{School of Artificial Intelligence and Data Engineering (SAIDE)} \\
        \textbf{Indian Institute of Technology Jodhpur} \\
        \textbf{\texttt{m25ai1132@iitj.ac.in}} \\}

\begin{document}
\maketitle

%  -  -  -  -  -  -  -  -  -  -  -  -  -  -  -  -  -  -  -  -  -  -  -  -  - -
\begin{abstract}
JEPA-family world models use a static predictor whose weights do not adapt
when test-time dynamics diverge from training.
We compare two mechanisms for incorporating accumulated experience into a
JEPA predictor under distribution shift: operand-side injection, where a
compressed experience representation is added as a residual to the
predictor's hidden state (\eijepa), and operator-side modulation, where the
same representation generates low-rank weight deltas via \lora applied to
the predictor's weights (\epmjepa).
On a pre-registered comparison (Moving MNIST, gravity shift), \epmjepa
(\dshift $= 0.7848 \pm 0.0078$, three seeds) differs from \eijepa
($0.8238$) by $\delta = 4.74\%$ - Outcome~C: a null result - by our stated
criterion, a valid outcome.
As a secondary, non-pre-registered observation, \epmjepa improves 1.90\%
over a no-memory baseline ($0.8000$), consistently across seeds, while
\eijepa underperforms the baseline, indicating the benefit is specific to
weight-level modulation.
Our primary contribution is a mechanism analysis: the \dshift trajectory
reflects three independent dynamical processes - buffer cycling, EMA target
drift, and an intrinsic \lora settling transient of $+0.021$ - rather than
convergence to equilibrium.
These findings motivate PEM-JEPA, a physics-grounded successor addressing
this dynamical-peak limitation.
\end{abstract}

% ============================================================================
\section{Introduction}
\label{sec:intro}

% Length target: ~0.75 pages.

LeCun's Joint-Embedding Predictive Architecture (JEPA) \citep{lecun2022jepa}
predicts future states in latent space rather than pixel space, sidestepping
the intractability of modelling irrelevant perceptual detail.
But the predictor is static: when world dynamics shift at test time - a
change in gravity, friction, or object behaviour - nothing in the
architecture signals that its predictions are now calibrated to a different
regime.
Memory mechanisms in transformers \citep{dai2019transformerxl,wu2022memorizing}
and differentiable memory models \citep{graves2014ntm,graves2016dnc} address
this on the \emph{operand side}: accumulated experience augments the
predictor's input, leaving its weights unchanged.
An alternative is \emph{operator-side} modulation - use experience to modify
the predictor's weights directly, so that the computation itself, not just its
input context, reflects what the model has seen.

We test this hypothesis with a controlled three-track comparison on Moving MNIST
\citep{srivastava2015lstm} with a gravity shift (base world: $0$~px/frame$^2$;
shift world: $0.5$~px/frame$^2$, encountered at inference after training on the
base world only).
Track~A is a \vjepa baseline with no memory.
Track~B (\eijepa) injects a compressed experience representation as a residual
into the predictor hidden state - the standard operand-side approach.
Track~C (\epmjepa) routes that same representation through a \lora modulator
that generates per-sample weight deltas
$\Delta W = U \cdot \mathrm{diag}(\delta) \cdot V^{\top}$ applied to the
predictor's linear layers - the operator-side alternative.

On the pre-registered test - $\delta = (D_B - D_C)/D_B = 4.74\%$, comparing
\epmjepa (Track~C, $0.7848 \pm 0.0078$) against \eijepa (Track~B, $0.8238$)
 - the result is Outcome~C (null result, $|\delta| < 5\%$), a valid
scientific result by our stated definition.
As a secondary, non-pre-registered observation, \epmjepa also improves
1.90\% over the no-memory baseline (Track~A, $0.8000$), consistent across
all three seeds.
\eijepa itself is \emph{worse} than Track~A, indicating that naive input
injection of experience does not help in this setting.
The primary contribution is a mechanism analysis showing that the performance
trajectory is governed by three independent dynamical processes - buffer
cycling, EMA drift, and a \lora settling transient of $+0.021$ in
$D_{\text{shift}}^{n=50}$ - rather than converging to a stable equilibrium.
These findings directly motivate PEM-JEPA, a physics-grounded successor
designed to address the dynamical-peak limitation identified here.
We report all results honestly: the pre-registered test yields a null
result; the mechanism characterisation, not the falsification verdict, is
the scientific payload.

% ============================================================================
\section{Related Work}
\label{sec:related}

% Length target: ~0.5 pages.

\paragraph{JEPA family.}
\citet{lecun2022jepa} proposed the Joint-Embedding Predictive Architecture
(JEPA) principle: a predictor should map between abstract latent representations
rather than reconstruct observations in pixel space, avoiding the intractable
modelling of irrelevant perceptual detail.
\citet{assran2023ijepa} instantiated this principle for static images (I-JEPA),
learning spatial structure via masked latent prediction without negative pairs.
\citet{bardes2024vjepa} extended the architecture to video (V-JEPA), predicting
the representations of masked spatiotemporal patches from unmasked context;
\citet{bardes2025vjepa2} further scaled this into V-JEPA~2, which demonstrated
that latent video prediction supports downstream understanding, prediction, and
planning.
EPM-JEPA builds directly on this family: we retain the JEPA prediction
objective and EMA target encoder, adding an experience-modulated predictor
that adapts to distribution shift online.

\paragraph{Memory-augmented neural networks.}
\citet{graves2014ntm} introduced the Neural Turing Machine, establishing
differentiable external memory with attention-based read and write heads;
\citet{graves2016dnc} refined this into the Differentiable Neural Computer,
adding dynamic memory allocation and link-based temporal addressing.
At the sequence-modelling level, Transformer-XL \citep{dai2019transformerxl}
extends context by caching segment-level hidden states across chunks, while
Memorizing Transformers \citep{wu2022memorizing} augment local attention with
$k$NN retrieval over a token-level external memory.
EPM-JEPA differs from all of these in that experience is not retrieved as an
additional input: it is compressed into a fixed-capacity buffer of
boundary-event embeddings and consumed via weight modulation, keeping the
predictor's interface unchanged.

\paragraph{Parameter-efficient adaptation.}
\citet{hu2021lora} showed that fine-tuning a large language model can be
reduced to learning a low-rank decomposition $\Delta W = UV^{\top}$ added to
frozen weights, achieving competitive performance at a fraction of the
parameter cost.
\citet{ha2017hypernetworks} proposed generating an entire network's weights
from a smaller conditioning hypernetwork; \citet{perez2018film} introduced
feature-wise linear modulation (FiLM), applying learned affine transforms to
intermediate representations conditioned on an external signal.
EPM-JEPA uses LoRA-style weight modulation, but the modulation vectors are
generated online from a live experience buffer rather than from a fixed
task embedding, making the weight delta a function of recent distributional
history rather than a static fine-tuning target.

\paragraph{SSL collapse prevention.}
\citet{bardes2022vicreg} introduced variance-invariance-covariance
regularisation (VICReg) to prevent dimensional collapse in non-contrastive
self-supervised learning, using a per-dimension variance hinge to enforce
output diversity.
\citet{grill2020byol} demonstrated that an online-target EMA network
suffices for SSL without negative pairs, provided the target weights lag the
online weights via exponential moving average.
EPM-JEPA inherits both: the EMA target encoder from BYOL and the VICReg
variance hinge.
A principal empirical finding of this work is that the two interact adversely
with \lora adaptation dynamics: the LoRA convergence window contracts the
output manifold, creating a structural tension between prediction optimality
and the variance hinge that cannot be resolved by tuning $\lambda$ alone.

% ============================================================================
\section{Method}
\label{sec:method}

% Length target: ~1.5 pages.

%  -  -  -  -  -  -  -  -  -  -  -  -  -  -  -  -  -  -  -  -  -  -  -  -  - -
\subsection{Problem Setup}
\label{sec:setup}

We consider the problem of multi-step latent prediction under distribution shift.
Let $x_t \in \mathbb{R}^{1 \times 64 \times 64}$ denote a grayscale video frame at time~$t$.
An encoder $E_\phi$ maps each frame to a latent vector $z_t = E_\phi(x_t) \in \mathbb{R}^{64}$
via four strided convolutional layers (channels $1 \to 16 \to 32 \to 64 \to 64$,
kernel size~4, stride~2) followed by a linear projection and layer normalisation
(${\approx}172\text{k}$ parameters; Listing~\ref{lst:encoder}).
A frozen EMA target encoder $\bar{E}_\phi$ - a deep copy of $E_\phi$ whose weights are
updated via a cosine-scheduled EMA with $\tau \in [0.996, 0.9999]$ - provides
stop-gradient prediction targets at three future horizons.
A predictor $P_\theta$ maps the current latent $z_t$ to a predicted future latent
$\hat{z}_{t+k}$ for horizons $k \in \{5, 10, 20\}$ frames ahead
(Listing~\ref{lst:predictor}).
The full training objective is given in Equation~\ref{eq:loss}
and hyperparameter choices are detailed in Section~\ref{sec:loss}.

%  -  -  -  -  -  -  -  -  -  -  -  -  -  -  -  -  -  -  -  -  -  -  -  -  - -
\subsection{Three-Track Architecture Comparison}
\label{sec:tracks}

All three tracks share the same encoder $E_\phi$, EMA target encoder $\bar{E}_\phi$,
predictor base $P_\theta$, experience encoding pipeline, and training objective.
They differ only in how accumulated experience, if any, is routed into the predictor.
Figure~\ref{fig:architecture} shows the full \epmjepa architecture.

\paragraph{Track A: Vanilla JEPA.}
\vjepa is the no-memory baseline.
The predictor operates directly on the current latent: $\hat{z}_{t+k} = P_\theta(z_t)$.
A frozen dummy $\mathrm{proj}_B \colon \mathbb{R}^{64} \to \mathbb{R}^{1024}$ is carried
but its output is discarded, preserving parameter-count comparability with Track~B.

\paragraph{Track B: EI-JEPA (Input Injection).}
\eijepa injects aggregated experience as a residual addition to the predictor's
first hidden state:
$h_1 = \mathrm{GELU}(W_1 z_t + b_1) + \mathrm{proj}_B(e_{\mathrm{agg}})$
and $\hat{z}_{t+k} = \mathrm{norm}(W_2 h_1 + b_2)$.
Here $\mathrm{proj}_B \colon \mathbb{R}^{64} \to \mathbb{R}^{1024}$ is a trainable
linear map that projects the aggregated experience vector into the predictor's hidden
dimension.
This operand-side injection modifies what the predictor operates on at its second layer
while leaving the predictor weights unchanged.

\paragraph{Track C: \epmjepa (LoRA Modulation).}
\epmjepa generates per-sample \lora weight deltas from the aggregated experience vector
and applies them to both predictor weight matrices~\citep{hu2021lora}, making this an
operator-side modulation (Listing~\ref{lst:lora}).
From $e_{\mathrm{agg}} \in \mathbb{R}^{64}$, two scale vectors are produced:
$\delta_\ell = \mathrm{Linear}_{\delta_\ell}(e_{\mathrm{agg}}) \in \mathbb{R}^r$
(no bias), with rank $r = 4$.
Each predictor weight matrix $W_\ell$ is then modulated as
$W_\ell^{\mathrm{eff}} = W_\ell + U_\ell\,\mathrm{diag}(\delta_\ell)\,V_\ell^\top$,
where $U_\ell, V_\ell$ are shared low-rank basis matrices
($U_1 \in \mathbb{R}^{1024 \times 4}$, $V_1 \in \mathbb{R}^{64 \times 4}$,
and transposed for layer~2).
Because the $\delta$ generators carry no bias term,
$e_{\mathrm{agg}} = \mathbf{0}$ (empty buffer) yields $\Delta W = \mathbf{0}$,
exactly recovering the Track~A identity path and ensuring graceful degradation before
any experience has been accumulated.

\paragraph{Memory Subsystem (B and C only).}
Tracks B and C share a four-component memory pipeline.
A \emph{boundary detector} monitors batch-mean surprisal
$s_t = \frac{1}{B}\sum_{b} \|z_t^b - \hat{z}_t^b\|_2$,
where $\hat{z}_t^b$ is produced by the base predictor (no \lora) to avoid circular
coupling, and maintains EMA running statistics $\mu_s$, $\sigma_s$ (momentum~0.99).
A boundary event fires when $s_t > \mu_s + \kappa\,\sigma_s$ after a 10-step
initialisation warmup; $\kappa = 1.5$ for Track~C and $\kappa = 2.0$ for Track~B
(Listing~\ref{lst:boundary}).
At each boundary event, the batch-mean latent transition $(z_{t-1}, z_t)$ is pushed
to a pre-allocated FIFO \emph{experience buffer} of capacity~256; both tensors are
detached before storage.
An \emph{experience encoder} - a 2-layer transformer~\citep{vaswani2017attention}
with $d = 64$, two attention heads, and feedforward dimension~128 - encodes each
stored pair as a 2-token sequence and mean-pools the output to $e_i \in \mathbb{R}^{64}$.
An \emph{attention aggregation} module then computes single-head soft attention over
all buffer entries:
$\alpha_i = \mathrm{softmax}(W_q(z_t)^\top W_k(e_i)\,/\sqrt{64})$,
$e_{\mathrm{agg}} = W_e\!\sum_i \alpha_i\,e_i \in \mathbb{R}^{64}$,
with the query derived from the current latent $z_t$.

%  -  -  -  -  -  -  -  -  -  -  -  -  -  -  -  -  -  -  -  -  -  -  -  -  - -
\subsection{Training Objective}
\label{sec:loss}

The total training loss decomposes into a prediction term and a variance regularisation term:

\begin{align}
  \mathcal{L} = {} & \underbrace{\mathbb{E}_t \!\!\sum_{k \in \{5,10,20\}}\!\!
    \big\|\, \mathrm{sg}[\bar{E}_\phi(x_{t+k})] - P_\theta(z_t) \big\|_2^2}_{\mathcal{L}_{\text{pred}}}
    \notag \\
    & \;+\; \lambda \underbrace{\mathbb{E}_d \max(0, \gamma - \sigma_d(P_\theta(z_t)))}_{\mathcal{L}_{\text{reg}}}
  \label{eq:loss}
\end{align}

$\mathcal{L}_{\mathrm{pred}}$ is the mean squared error between each predicted latent
and the corresponding stop-gradient EMA target, averaged over the three horizons
$k \in \{5, 10, 20\}$ (Listing~\ref{lst:loss}).
$\mathcal{L}_{\mathrm{reg}}$ is the variance component of VICReg~\citep{bardes2022vicreg}:
a per-dimension hinge that penalises the standard deviation of predictor output falling
below $\gamma$, without covariance or invariance terms.
This minimal regulariser prevents latent collapse while imposing no constraint beyond
a soft lower bound on output diversity.

Hyperparameters $\lambda = 0.05$ and $\gamma = 0.75$ were selected during Phase~1
sequential tuning (Section~\ref{sec:phase1}) and held fixed across all subsequent
experiments.
The batch size is~64; the learning rate follows cosine annealing with warm
restarts~\citep{loshchilov2017sgdr}, with initial rates of $3 \times 10^{-3}$ for
Tracks~A and~C and $2 \times 10^{-3}$ for Track~B.
All trainable parameters are updated by a single AdamW~\citep{loshchilov2019adamw}
optimiser group.

%  -  -  -  -  -  -  -  -  -  -  -  -  -  -  -  -  -  -  -  -  -  -  -  -  - -
\clearpage
\subsection{Architecture Diagram}

\begin{strip}
  \centering
  \includegraphics[width=\textwidth]{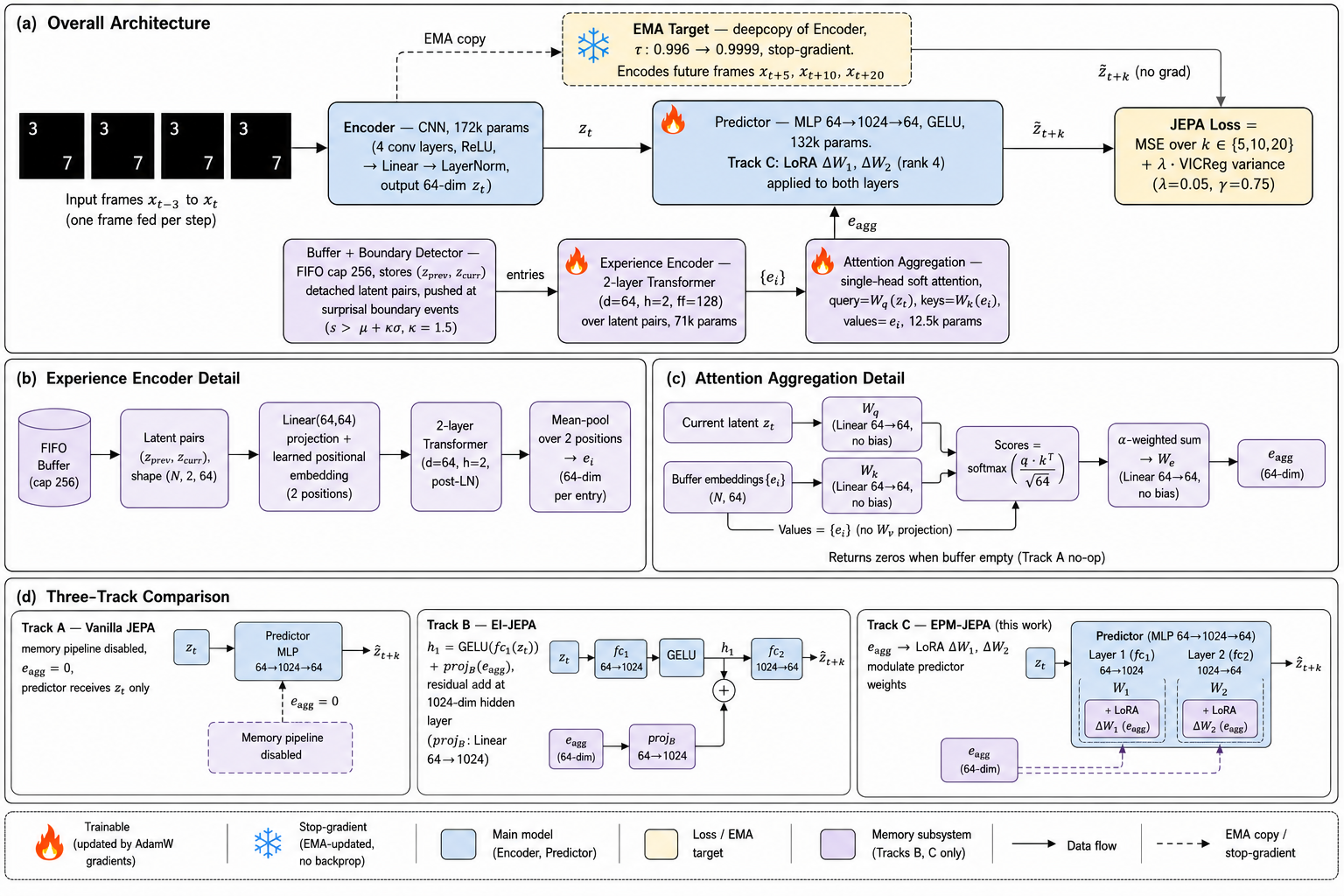}
  \captionof{figure}{\textbf{\epmjepa architecture overview.}
  \textbf{(a)} Overall architecture: the encoder maps each input frame to a
  latent state $z_t$; the EMA target encoder provides stop-gradient
  prediction targets for the JEPA loss; the predictor (with \lora weight
  modulation in Track~C) produces $\hat{z}_{t+k}$ for $k \in \{5,10,20\}$;
  the memory subsystem (boundary detector, experience buffer, experience
  encoder, attention aggregation) encodes accumulated experience into
  $e_{\mathrm{agg}}$.
  \textbf{(b)} Experience encoder detail: each buffered transition pair is
  encoded by a 2-layer transformer and mean-pooled to $e_i \in \mathbb{R}^{64}$.
  \textbf{(c)} Attention aggregation detail: the current latent $z_t$ attends
  over buffer entries to produce $e_{\mathrm{agg}}$.
  \textbf{(d)} Three-track comparison: Track~A (\vjepa, no memory), Track~B
  (\eijepa, residual injection), and Track~C (\epmjepa, \lora weight
  modulation) share the encoder, EMA target, and predictor base, differing
  only in how $e_{\mathrm{agg}}$ is consumed.}
  \label{fig:architecture}
\end{strip}

% ============================================================================
\section{Experiments}
\label{sec:experiments}

% Length target: ~2.5 pages including figures.

%  -  -  -  -  -  -  -  -  -  -  -  -  -  -  -  -  -  -  -  -  -  -  -  -  - -
\subsection{Setup}
\label{sec:setup-exp}

We use a synthetic Moving MNIST variant~\citep{srivastava2015lstm} as a controlled
testbed for distribution shift.
Each sequence contains 21 frames ($t = 0, \ldots, 20$) of two MNIST digits moving on a
$64 \times 64$ grayscale canvas, with initial per-digit velocity sampled uniformly from
$[2, 4]$~px/frame.
Two worlds are generated from the same digit pool: a \emph{base world} (gravity $= 0.0$,
linear motion) of 10,000 sequences split 8,000/1,000/1,000 (train/val/test), and a
\emph{shift world} (downward gravity $= 0.5$~px/frame$^2$, Euler-integrated per frame)
of 1,000 sequences split 800/100/100.
All models are trained exclusively on base-world training sequences; the shift world
is held out for evaluation only.

Experiments run on a single GTX~1050~Ti (4~GB VRAM, Pascal compute~6.1) in fp32
precision without AMP or \texttt{torch.compile}.
Phase~1 and grid runs use a 600~s time budget; extended multi-seed runs use 3,600~s,
with throughput declining from ${\approx}17.6$ to ${\approx}9.6$ steps/s due to thermal
throttling on longer runs.
The primary metric is \dshift: mean normalised prediction error
$\mathbb{E}[\|z_{t+k} - \hat{z}_{t+k}\|\,/\,\|z_{t+k} - z_t\|]$
averaged over $k \in \{5, 10, 20\}$, evaluated on the shift-world test set at $N = 50$
accumulated shift-world experiences; samples where $\|z_{t+k} - z_t\| < 10^{-3}$ are
excluded to guard against degenerate denominators.
A score of 1.0 corresponds to trivial constant prediction; lower is better.
The secondary metric \embedstd is the mean per-dimension standard deviation of predictor
output over an evaluation batch, used to detect representation collapse.
We pre-registered a falsification criterion comparing \epmjepa (Track~C)
against \eijepa (Track~B): $\delta = (D_B - D_C)/D_B$, with four outcome
classes - $\delta \geq 0.20$ (Outcome~A, strong confirmation),
$0.05 \leq \delta < 0.20$ (Outcome~B, partial confirmation),
$|\delta| < 0.05$ (Outcome~C, null result), and $\delta < -0.05$
(Outcome~D, falsified).
Outcome~C is explicitly designated a valid scientific result under this
definition.
The outcome is reported in Section~\ref{sec:main}.

%  -  -  -  -  -  -  -  -  -  -  -  -  -  -  -  -  -  -  -  -  -  -  -  -  - -
\subsection{Phase 1: Sequential Hyperparameter Tuning}
\label{sec:phase1}

Phase~1 consists of 25 sequential single-parameter experiments on Track~A (seed~42,
600~s budget each), locking the best value found before sweeping the next hyperparameter.
\dshift improved from 1.8156 at random initialisation to 0.8000 at the ratchet's
conclusion - a 55.9\% reduction; the full experiment log is in
Appendix~\ref{app:phase1}.
Three findings shaped the locked configuration used for all subsequent experiments.
First~(F2), learning rate and $\lambda$ interact: LR~$= 3 \times 10^{-3}$ was the worst
performer at $\lambda = 1.0$ but the best at $\lambda = 0.05$, motivating the
$\lambda = 0.05$ lock and a Phase~2 grid search for Track~B's learning rate.
Second~(F3), $\gamma = 0.75$ is the universal optimum across all tracks; lower values
allow latent collapse and higher values over-penalise output variance.
Third~(F4), HIDDEN\_DIM~$= 1024$ outperforms 768 and 512, establishing predictor
capacity as a meaningful factor independent of the experience mechanism.

%  -  -  -  -  -  -  -  -  -  -  -  -  -  -  -  -  -  -  -  -  -  -  -  -  - -
\subsection{Main Result: Three-Track Comparison}
\label{sec:main}

Table~\ref{tab:main} summarises the primary result: mean \dshift at peak step across
three seeds per track, bootstrapped 95\% confidence intervals, and \embedstd.

\begin{table*}[t]
  \centering
  \caption{\textbf{Three-track comparison on Moving MNIST with gravity shift.}
  Track C (\epmjepa) achieves the best \dshift at peak (step 10000) across
  three seeds. The pre-registered comparison (Track~C vs Track~B) yields
  Outcome~C (null result, $\delta = 4.74\%$); as a secondary,
  non-pre-registered observation, \epmjepa is 1.90\% below Track~A's best
  (0.8000), with non-overlapping bootstrap 95\% CIs.}
  \label{tab:main}
  \small
  \begin{tabular}{lcccc}
    \toprule
    Track & System & \dshift (3-seed mean) & 95\% CI & \embedstd \\
    \midrule
    A & \vjepa & 0.8089 & [0.8000, 0.8159] & 0.5321 \\
    B & \eijepa & 0.8238 & [0.8217, 0.8249] & 0.5324 \\
    C & \epmjepa & \textbf{0.7848} & [0.7776, 0.7931] & 0.4123 \\
    \bottomrule
  \end{tabular}
\end{table*}

On the pre-registered comparison - $\delta = (D_B - D_C)/D_B = 4.74\%$,
where \eijepa (Track~B) achieves mean \dshift $= 0.8238$ and \epmjepa
(Track~C) achieves mean \dshift $= 0.7848 \pm 0.0078$ across three seeds at
step~10,000 - the result is Outcome~C (null result, $|\delta| < 5\%$), a
valid scientific result under the criterion defined in
Section~\ref{sec:setup}.
As a secondary, non-pre-registered observation, \epmjepa also improves
1.90\% over \vjepa's (Track~A) best single-seed result ($0.8000$) and
2.98\% over Track~A's three-run mean ($0.8089$).
We report both results directly: the pre-registered test is null, the
secondary comparison is modest but reproduces across all three seeds, and we
treat the mechanism characterisation in Section~\ref{sec:mechanism} as the
primary scientific contribution.

At the per-seed level, all three Track~C seeds (0.7776, 0.7838, 0.7931)
individually beat Track~A's single best result (0.8000) - a secondary
observation, not part of the pre-registered test.
Bootstrap 95\% CIs over $N = 3$ canonical runs - [0.7776, 0.7931] for Track~C
and [0.8000, 0.8159] for Track~A - do not overlap, suggesting this secondary
gap is consistent even at this small~$N$.
Track~B (\eijepa) underperforms both Track~A and Track~C with mean $0.8238$;
operand-side injection of experience into the predictor's hidden state does
not help in this setting.
This points to the \emph{pathway} of experience application - weight
modulation versus input modulation - as the determining factor, not merely
the presence of a memory mechanism.

Two caveats accompany Table~\ref{tab:main}.
First, Track~A seeds~43 and~44 were thermally truncated at steps~6404 and~6868
respectively; their \dshift values (0.8159, 0.8108) may be mildly pessimistic relative
to a full-budget run, meaning the true Track~A mean is at most equal to its reported
value - Track~C's advantage is therefore conservative, not inflated.
Second, Track~C \embedstd ranges 0.4123-0.4253 across seeds at peak, below the 0.5
safety threshold; this structural tension between low \dshift and low \embedstd is
examined in Section~\ref{sec:svd}.

\begin{figure*}[t]
  \centering
  \includegraphics[width=0.7\textwidth]{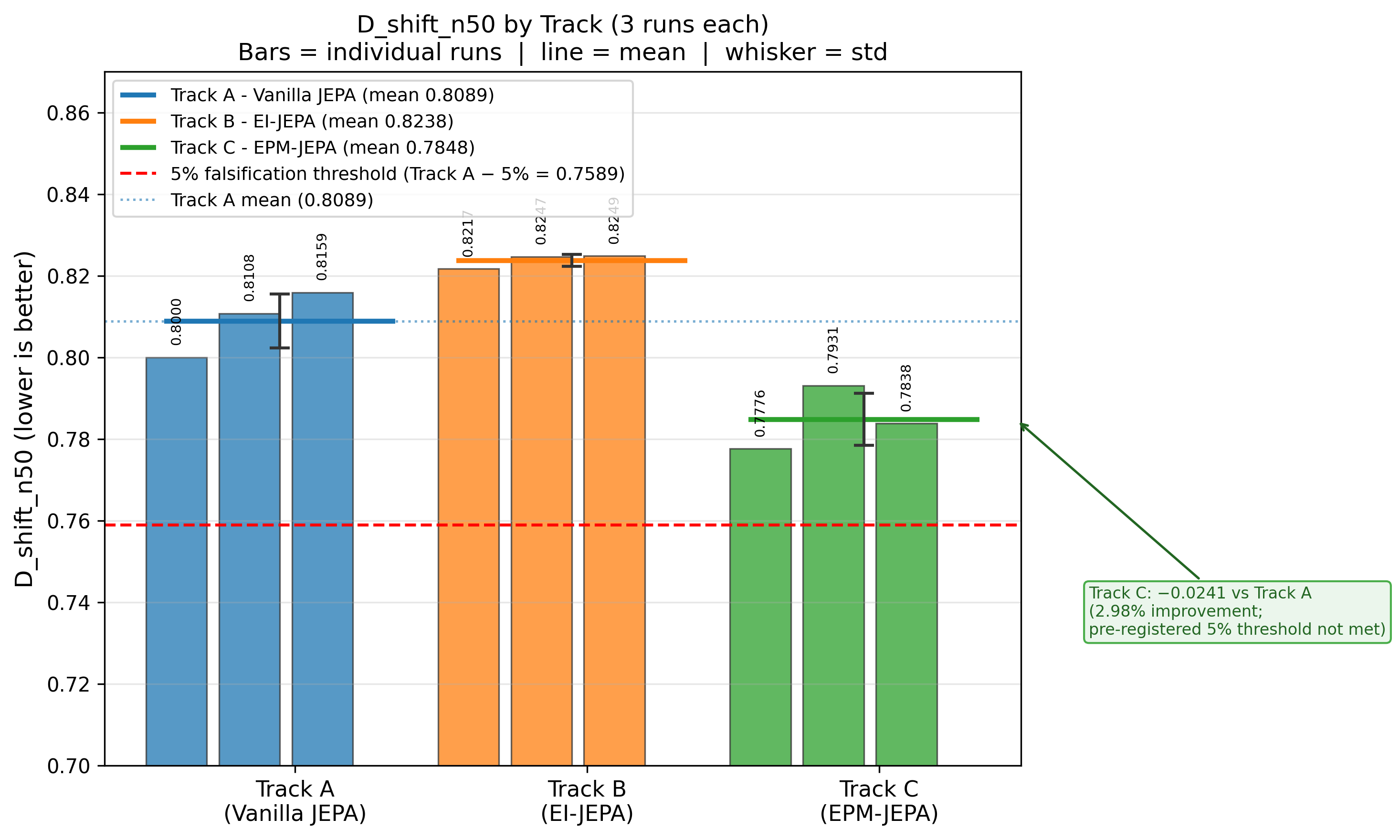}
  \caption{\textbf{Multi-seed \dshift at peak step (step $\approx$10000).}
  Each bar is one canonical run; horizontal line and whisker show mean $\pm$
  std. The pre-registered comparison (Track~C vs Track~B) yields Outcome~C
  (null result, $\delta = 4.74\%$, $|\delta| < 5\%$). As a secondary,
  non-pre-registered observation, all three Track~C seeds achieve lower
  \dshift than Track~A's best single-seed result (1.90\% advantage),
  consistent across seeds. Track A seeds 43/44 thermally truncated at
  6400/6900 steps and reported with this caveat.}
  \label{fig:multiseed}
\end{figure*}

\begin{figure*}[t]
  \centering
  \begin{subfigure}{0.48\textwidth}
    \includegraphics[width=\textwidth]{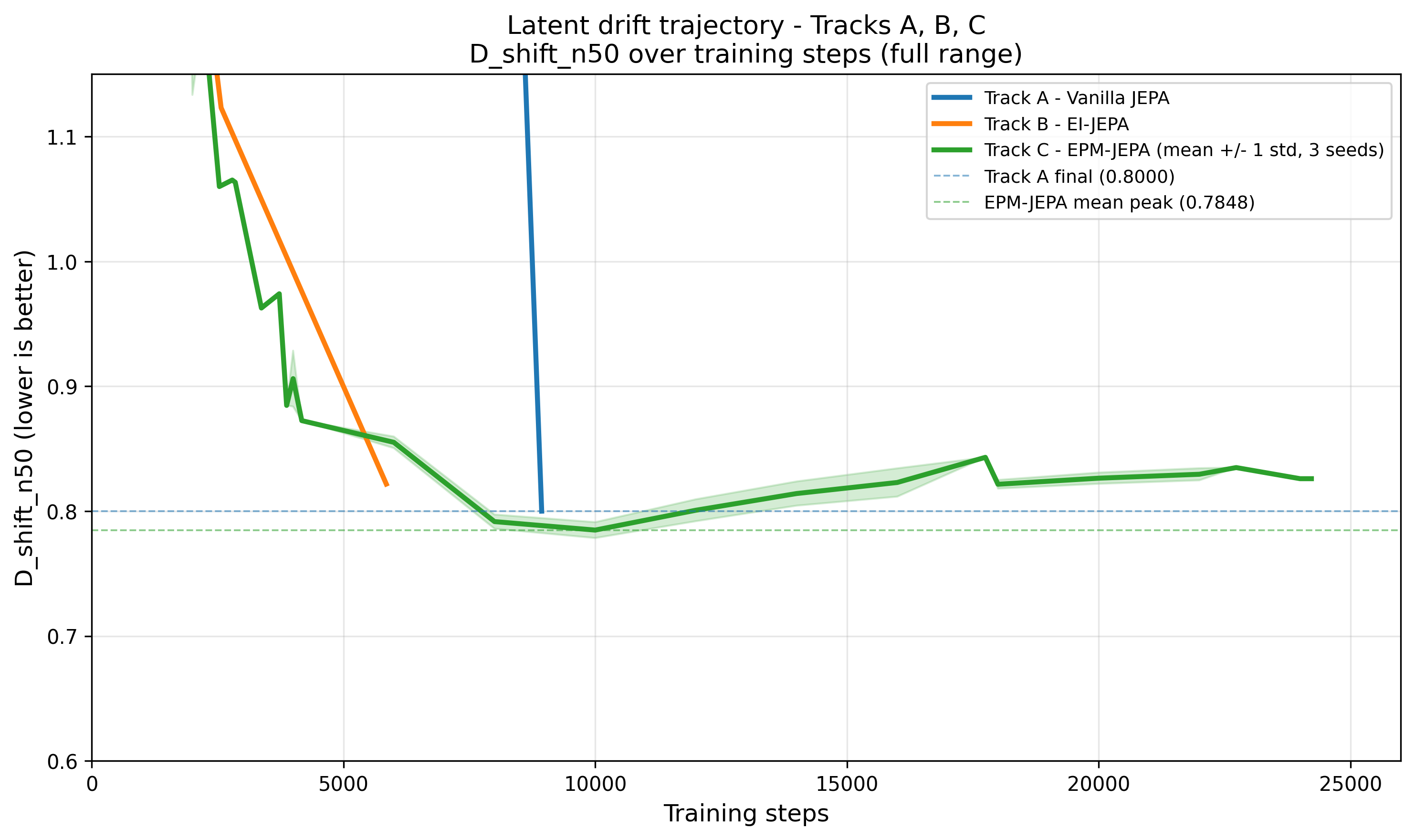}
    \caption{Full training range (\dshift 0.60-1.15). Tracks A and B descend
    from above 1.0; Track C (mean $\pm$ 1 std, 3 seeds, shaded) converges
    to a lower minimum earlier in training.}
  \end{subfigure}
  \hfill
  \begin{subfigure}{0.48\textwidth}
    \includegraphics[width=\textwidth]{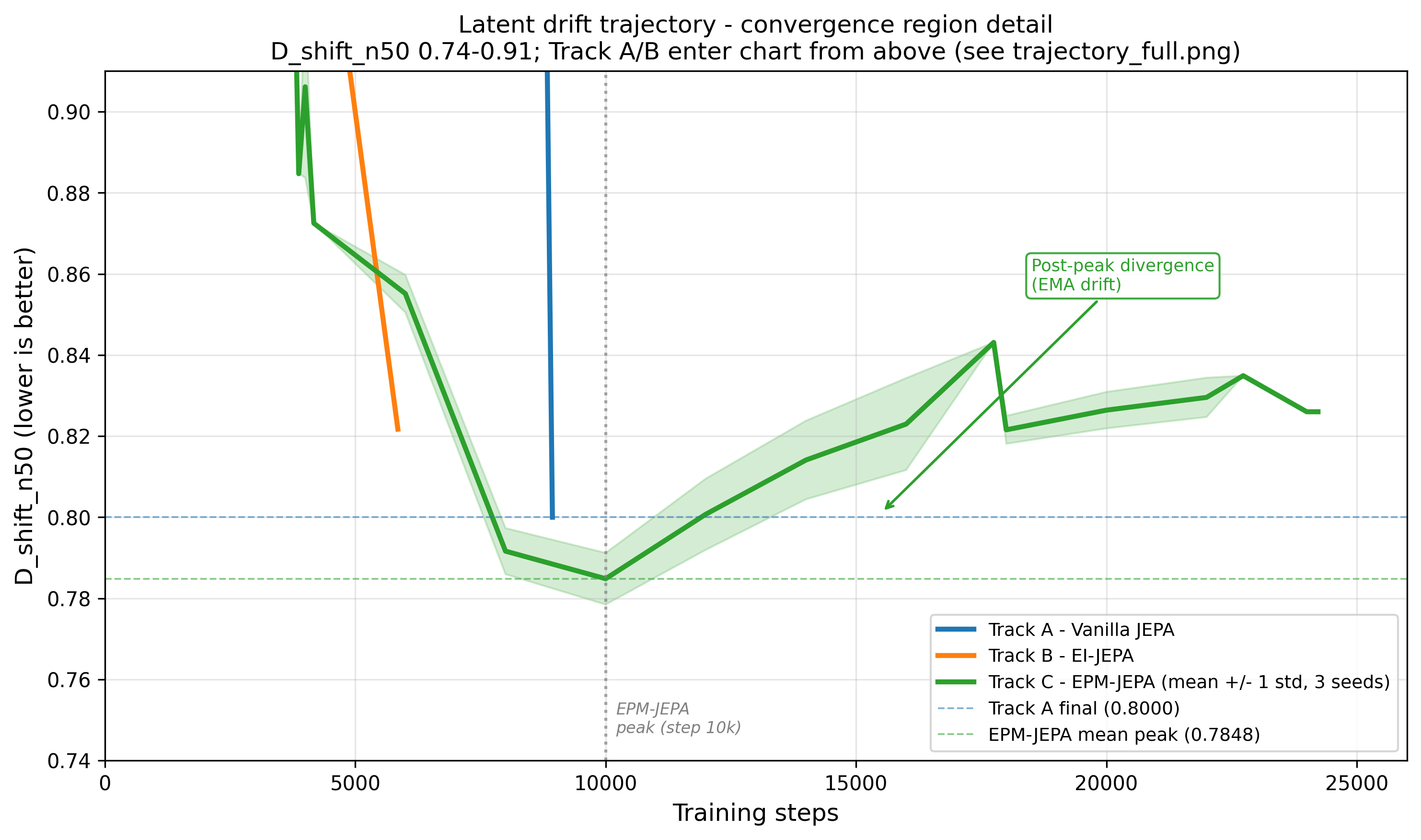}
    \caption{Convergence region (0.7400-0.9100). Track C peaks at step
    $\approx$10000 (dashed vertical) then diverges (EMA drift, annotated).
    Tracks A and B enter the visible range from above. See
    Section~\ref{sec:mechanism} for ablation isolating the divergence cause.}
  \end{subfigure}
  \caption{\textbf{\dshift training trajectories - full range and convergence
  detail.} Dashed horizontal lines: Track A final (0.8000) and \epmjepa mean
  peak (0.7848). Shaded band is $\pm 1$ std across 3 seeds (Track C only).}
  \label{fig:trajectory}
\end{figure*}

Figure~\ref{fig:multiseed} shows \dshift at peak step for each canonical run; all three
Track~C seeds fall below Track~A's best, and the Track~A bars for seeds~43 and~44 carry
the thermal-truncation caveat described above.
Figure~\ref{fig:trajectory} shows the training dynamics: the full-range panel shows
Tracks~A and~B descending to the 0.80 region from above 1.0, while Track~C (mean
$\pm 1$ std across three seeds, shaded) converges to a lower floor earlier in training;
the zoomed panel isolates the convergence region and annotates the post-peak divergence
whose mechanism is analysed in Section~\ref{sec:mechanism}.

%  -  -  -  -  -  -  -  -  -  -  -  -  -  -  -  -  -  -  -  -  -  -  -  -  - -
\subsection{Mechanism Analysis: Three Dynamical Mechanisms}
\label{sec:mechanism}

% This is the paper's primary contribution. Aim for ~1 page.

\begin{figure*}[t]
  \centering
  \includegraphics[width=0.7\textwidth]{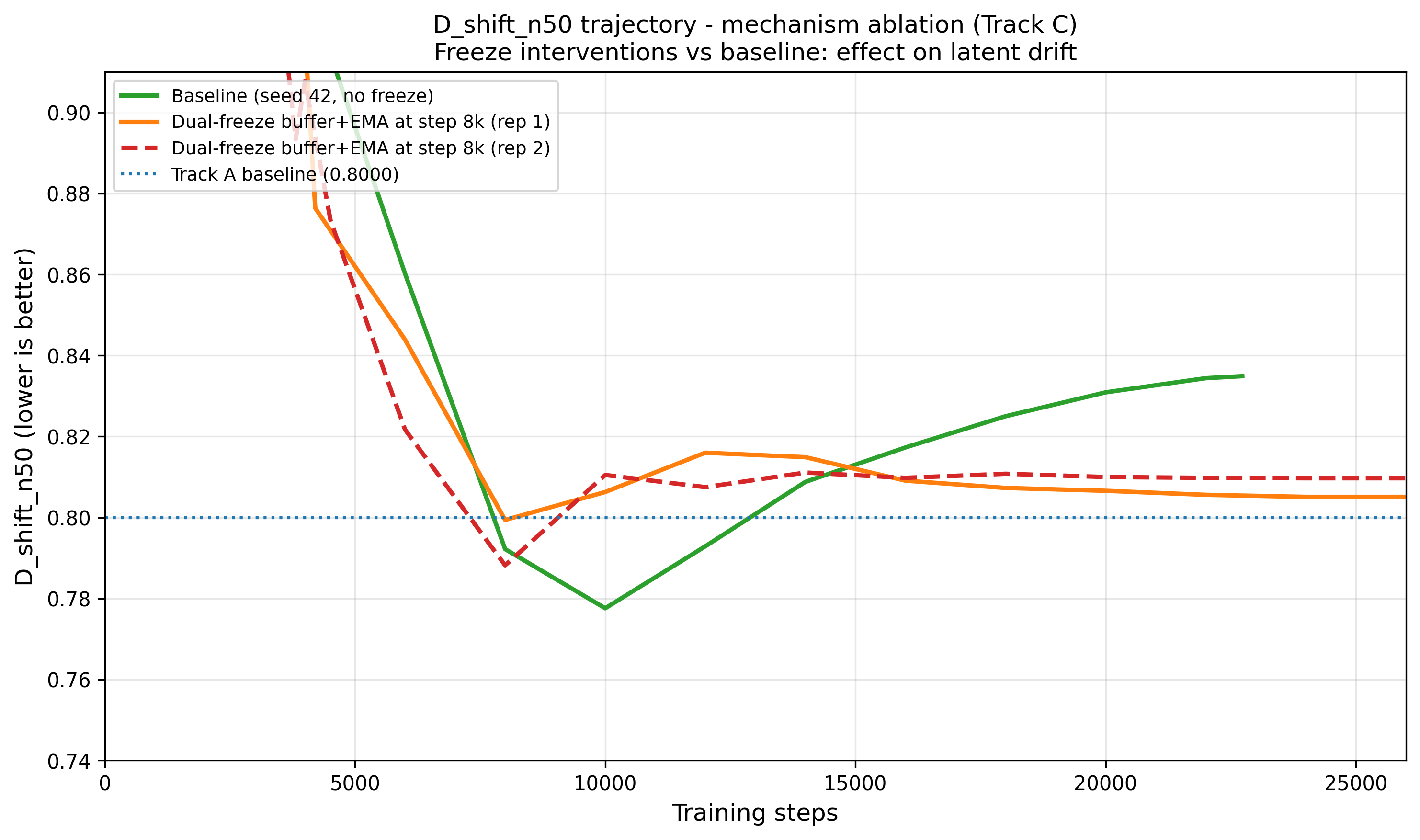}
  \caption{\textbf{\dshift trajectories across mechanism-ablation experiments.}
  Baseline seed 42 (green) peaks at step $\approx$10000 then diverges to
  0.8349. Dual-freeze (buffer + EMA at step $\approx$8000, orange/red) arrests
  the divergence and reveals an intrinsic +0.0210 settling transient that
  appears in both replications regardless of timing. Horizontal dotted line:
  Track A baseline (0.8000). See Table~\ref{tab:mechanism} for peak, final,
  and settling $\Delta$ per condition.}
  \label{fig:trajectories}
\end{figure*}

\epmjepa exhibits a characteristic trajectory across all three no-freeze seeds:
\dshift improves monotonically from initialisation to a peak near step~$10{,}000$,
then degrades through the remainder of training to final values of 0.8260-0.8431
(Figure~\ref{fig:trajectories}).
This post-peak divergence is reproducible and not an artefact of a single run.
To characterise the underlying dynamics, we run targeted ablation experiments that
jointly freeze the experience buffer and EMA target encoder at prescribed steps and
continue training; results are summarised in Table~\ref{tab:mechanism}.
We identify three dynamical mechanisms contributing to this trajectory.

\begin{table*}[t]
  \centering
  \caption{\textbf{Mechanism ablation summary.} All freeze conditions jointly
  freeze the experience buffer and EMA target at the specified step. Peak and
  final \dshift, plus post-freeze settling $\Delta = \text{Final} - \text{Peak}$.
  Rep~2 and the late-freeze run show consistent settling ($+0.0215$ vs $+0.0209$)
  across a 2,000-step difference in freeze timing, consistent with an intrinsic
  \lora settling transient; rep~1 shows a smaller transient ($+0.0057$),
  discussed in the text.}
  \label{tab:mechanism}
  \small
  \begin{tabular}{lccccc}
    \toprule
    Experiment & Buffer freeze & EMA freeze & Peak \dshift & Final \dshift & Settling $\Delta$ \\
    \midrule
    Baseline (seed 42) & - & - & 0.7776 & 0.8349 & +0.0573 \\
    Dual-freeze (rep 1) & step 8000 & step 8000 & 0.7994 & 0.8051 & +0.0057 \\
    Dual-freeze (rep 2) & step 8000 & step 8000 & 0.7882 & 0.8097 & +0.0215 \\
    Dual-freeze (late) & step 10000 & step 10000 & 0.7996 & 0.8205 & +0.0209 \\
    \bottomrule
  \end{tabular}
\end{table*}

\paragraph{Mechanism 1: Buffer cycling affects peak formation.}
The experience buffer reaches capacity at approximately step~3700-4200 (256 entries
at the observed boundary-event rate), after which each new event evicts the oldest
entry and LoRA continues adapting to a continuously cycling experience distribution.
The convergence peak at step~$\approx 10{,}000$ emerges when the LoRA weight
configuration has settled into a distribution consistent with the current cycling prior,
roughly 2-3 full buffer turnovers after the buffer first fills.
The two step-8000 dual-freeze replications achieve different \dshift values at the
freeze point (rep~1: 0.7994, rep~2: 0.7882), consistent with run-dependent LoRA
adaptation state at that moment; a clean buffer-only ablation (buffer frozen, EMA
unfrozen) was not conducted and is a planned follow-up for isolating this mechanism
directly.

\paragraph{Mechanism 2: EMA target drift causes post-peak divergence.}
Without any freeze, the baseline diverges from its peak of 0.7776 to a final
\dshift of 0.8349 ($\Delta = +0.0573$) over roughly 12,000 further training steps.
All three dual-freeze conditions substantially arrest this divergence: final \dshift
of 0.8051 and 0.8097 for the step-8000 replications, and 0.8205 for the step-10000
run.
Because freezing the EMA target encoder is the common element across all arrested
conditions, this is consistent with slow EMA target drift as the primary driver of
post-peak divergence: after LoRA settles, the EMA target encoder continues its
momentum-weighted updates (approaching $\tau \to 0.9999$ under the cosine schedule),
gradually shifting the prediction target and pulling \dshift upward.
Freezing the EMA arrests this drift and stabilises the training signal.

\paragraph{Mechanism 3: \lora settling is intrinsic and freeze-invariant.}
After both freezes fire, the \lora $\delta$ generators continue updating under the
cosine-decaying learning rate for approximately 2,000 further steps, producing a
residual rise in \dshift before the model reaches its stable post-freeze configuration.
The settling magnitude is consistent between dual-freeze rep~2 ($\Delta = +0.0215$,
frozen at step~8000) and the late dual-freeze run ($\Delta = +0.0209$, frozen at
step~10,000) - an agreement within 0.0006 across a 2,000-step difference in freeze
timing - consistent with this transient being intrinsic to \lora dynamics under
decaying learning rate rather than a function of the model state at the freeze point.
Rep~1 exhibits a smaller transient ($\Delta = +0.0057$), suggesting that \lora had
largely settled by step~8000 in that replication.
The peak is therefore a transient dynamical configuration: any intervention that
freezes external signals reveals a residual \lora settling phase before the model
stabilises, and the peak cannot be captured by halting training at step~10,000.

%  -  -  -  -  -  -  -  -  -  -  -  -  -  -  -  -  -  -  -  -  -  -  -  -  - -
\subsection{Hyperparameter Sensitivity}
\label{sec:hparam}

Table~\ref{tab:lambda} shows \dshift and \embedstd at peak step for three values of
$\lambda$ on Track~C (seed~42, 60~min budget).

\begin{table}[t]
  \centering
  \caption{\textbf{\lora regularization sweep (Track C, seed 42, 60 min).}
  $\lambda = 0.05$ is uniquely optimal; both higher values degrade \dshift.}
  \label{tab:lambda}
  \small
  \begin{tabular}{lccc}
    \toprule
    $\lambda$ & Peak \dshift & Final \dshift & \embedstd at peak \\
    \midrule
    0.05 & \textbf{0.7776} & 0.8349 & 0.4123 \\
    0.10 & 0.8139 & 0.8619 & 0.4661 \\
    0.15 & 0.8009 & 0.8595 & 0.4760 \\
    \bottomrule
  \end{tabular}
\end{table}

$\lambda = 0.05$ is uniquely optimal: peak \dshift $= 0.7776$, with convergence at
step~10,000.
Both higher values degrade prediction performance; $\lambda = 0.10$ peaks at step~8,000
with \dshift $= 0.8139$, and $\lambda = 0.15$ also peaks at step~8,000 with
\dshift $= 0.8009$.
Heavier regularisation shifts the convergence peak 2,000 steps earlier while worsening
performance, suggesting the variance hinge interferes with \lora's adaptation dynamics
before the natural convergence point is reached.
Higher $\lambda$ raises \embedstd monotonically (0.4123 $\to$ 0.4661 $\to$ 0.4760),
partially recovering output diversity, but no value in the sweep simultaneously achieves
\dshift~$< 0.80$ and \embedstd~$\geq 0.5$ at peak.
This trade-off is structural: the \lora weight configuration that achieves the best
prediction contracts the output manifold, while the VICReg variance hinge pushes in the
opposite direction; increasing $\lambda$ resolves the collapse concern only by sacrificing
the prediction gain that motivates Track~C in the first place.

%  -  -  -  -  -  -  -  -  -  -  -  -  -  -  -  -  -  -  -  -  -  -  -  -  - -
\subsection{Representation Health: Embedding Diversity}
\label{sec:svd}

\begin{figure*}[t]
  \centering
  \includegraphics[width=0.7\textwidth]{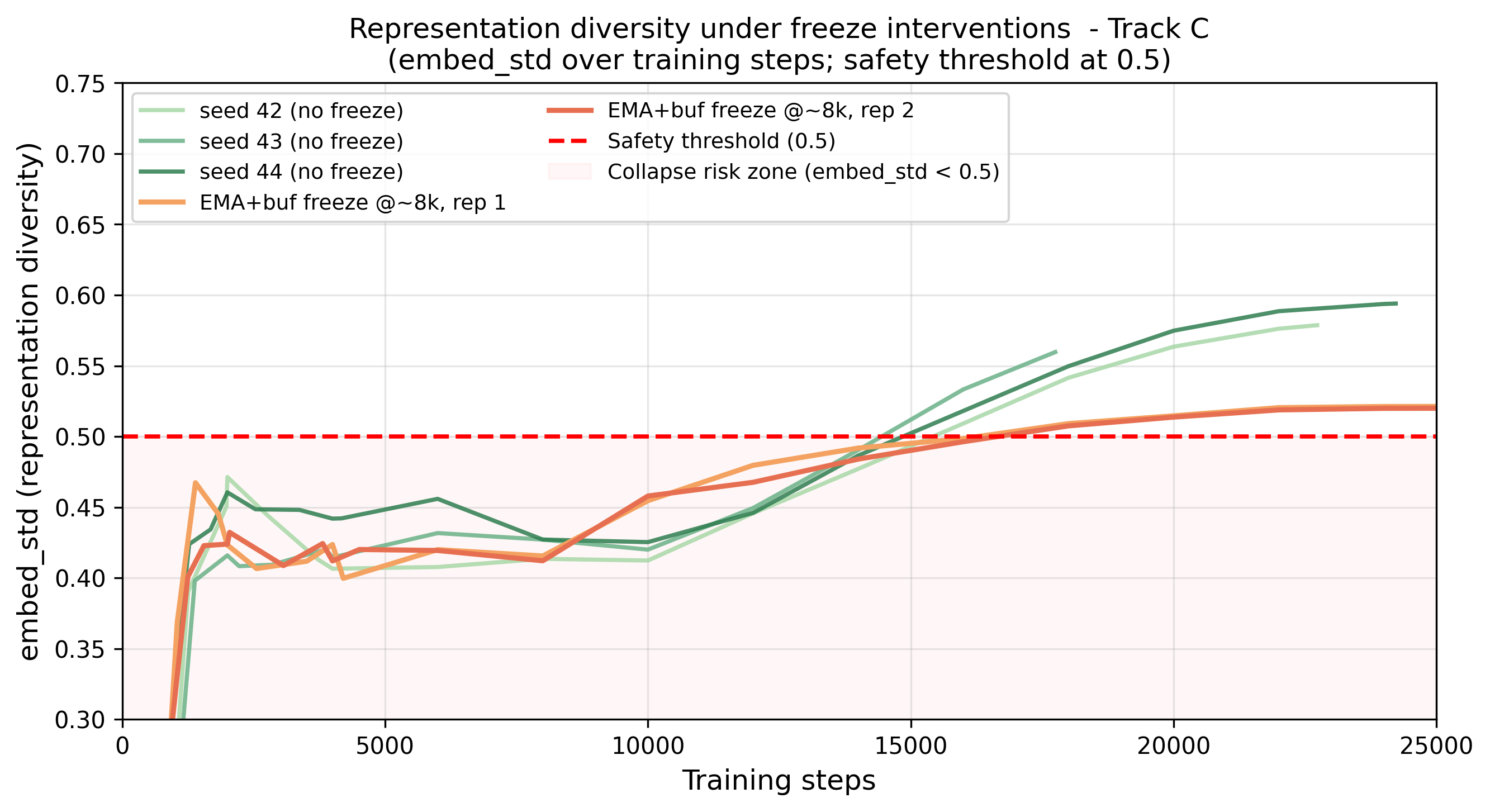}
  \caption{\textbf{\embedstd over training steps for Track C.}
  No-freeze seeds (42/43/44, green shades) show \embedstd declining
  through training and falling below the 0.5 safety threshold near peak.
  Freeze runs (orange) stabilize \embedstd above 0.5 after the freeze step.
  Shaded region below 0.5 marks collapse risk zone. X-axis limited to
  25000 steps; seed 44 reaches step 24225.}
  \label{fig:embedstd}
\end{figure*}

All three no-freeze Track~C seeds reach \embedstd $\in \{0.4123, 0.4200,
0.4253\}$ at the prediction peak (step~10,000) - below the 0.5 safety
threshold visible in Figure~\ref{fig:embedstd}.
This is not classical representation collapse: the VICReg variance hinge
($\gamma = 0.75$) remains active throughout training, and no per-dimension
constant-output behaviour was observed.
Rather, the compression is a transient artefact of \lora settling - as the
experience aggregate \(e_{\mathrm{agg}}\) stabilises during the convergence
window, the modulated predictor outputs converge to a narrower region of the
latent space.
Crucially, the dip is self-correcting: all three seeds recover above the
threshold post-peak, ending at \embedstd $\in \{0.5787, 0.5597, 0.5940\}$,
confirming that the manifold contraction is coupled to the \lora adaptation
phase rather than representing a permanent structural failure.

The freeze runs corroborate this interpretation.
Arresting EMA and buffer updates near the peak step keeps \embedstd above 0.5
throughout the post-freeze period (final values $0.5213$, $0.5200$,
$0.5467$), consistent with the view that EMA drift - not \lora per se - 
sustains the contraction beyond step~10,000.
However, this is achieved at the cost of the $+0.021$ \dshift settling
transient documented in Section~\ref{sec:mechanism}: the same freeze that
stabilises representation health also locks in the LoRA-settling penalty.
SVD spectral analysis (effective rank, singular value distribution) was planned
to characterise the degree of manifold compression quantitatively, but requires
a saved model checkpoint; canonical training runs did not persist weights
(see Section~\ref{sec:discussion}), and this analysis is deferred to the
RunPod scale-up experiment.

%  -  -  -  -  -  -  -  -  -  -  -  -  -  -  -  -  -  -  -  -  -  -  -  -  - -
\subsection{Buffer Attention: Qualitative Analysis}
\label{sec:attention}

We planned to visualise how \epmjepa attends over experience-buffer entries
during shift-world inference - specifically, whether recent boundary events
receive disproportionate attention weight relative to older entries.
However, canonical training runs operated under a wall-clock budget with no
end-of-run checkpoint save; weights were written only on thermal-pause signal
(see Appendix~\ref{app:compute}), and no \texttt{.pt} files were retained for
the reported runs.
Attention analysis is therefore deferred to the planned RunPod scale-up
experiment, where checkpoint saving will be enabled from the first step.

% ============================================================================
\section{Discussion}
\label{sec:discussion}

% Length target: ~0.75 pages.

\subsection{What we learned}

On the pre-registered test - $\delta = (D_B - D_C)/D_B = 4.74\%$, comparing
\epmjepa (Track~C, $0.7848 \pm 0.0078$ across three seeds at the convergence
peak, step~10,000) against \eijepa (Track~B, mean $0.8238$) - the result is
Outcome~C: a null result ($|\delta| < 5\%$) that is, by our own stated
definition, a valid scientific result.
As a secondary, non-pre-registered observation, \epmjepa improves 1.90\% over
the no-memory baseline (Track~A best, $0.8000$); bootstrap confidence
intervals ($[0.7776, 0.7931]$ for Track~C; $[0.8000, 0.8159]$ for Track~A) do
not overlap, suggesting this secondary direction is consistent even if the
magnitude is modest.
Notably, \eijepa itself performs worse than Track~A (mean $0.8238$ vs
$0.8089$), demonstrating that naive input injection does not help in this
setting - the benefit, where it exists, is specific to weight-level
modulation via \lora: an experience pathway that influences predictor
computation rather than predictor input.

The primary scientific contribution of this work is the mechanism analysis.
The \dshift trajectory is not converging to a stable equilibrium: it is the
outcome of three independent dynamical processes operating on different
timescales.
Buffer cycling governs peak formation as the finite buffer saturates and begins
cycling older experiences out; EMA drift governs post-peak divergence as the
slowly moving target encoder continues to pull the prediction target away after
\lora has settled; and the LoRA settling transient ($+0.021$ in \dshift,
regardless of freeze timing) is an intrinsic property of the low-rank
adaptation dynamics.
Freeze-based interventions can arrest EMA drift and thereby stabilise long-run
performance, but they cannot eliminate the LoRA settling transient - pointing
to the modulation architecture itself, rather than the training schedule, as
the lever for future improvement.

A secondary finding concerns the structural tension between prediction
optimality and representation diversity.
All three Track~C seeds reach \embedstd $\in \{0.4123, 0.4200, 0.4253\}$ at
peak - below the 0.5 safety threshold - precisely when \dshift is best.
Increasing the VICReg weight $\lambda$ raises \embedstd but degrades \dshift
and shifts convergence 2,000 steps earlier; no value in the sweep
simultaneously achieves \dshift~$< 0.80$ and \embedstd~$\geq 0.5$ at peak.
The compression is a transient artefact of \lora adaptation (all seeds recover
to 0.56-0.59 post-peak) and does not constitute representation collapse, but
the trade-off is structural and warrants architectural rather than
regularisation-based resolution.

\subsection{Limitations}

Several limitations qualify the conclusions above.
First, all experiments use a single dataset (Moving MNIST) and a single
distribution shift type (gravity magnitude), so it is unknown whether the
findings generalise to richer visual domains or qualitatively different shifts
such as object identity or scene illumination.
Second, the pre-registered test (Track~C vs Track~B) produced Outcome~C - 
a null result, $\delta = 4.74\%$, $|\delta| < 5\%$ - which is, by our own
stated definition, a valid scientific result; the secondary comparison
against Track~A (1.90\%) was not pre-registered and should be interpreted
accordingly.
Third, Track~A and Track~B results used for comparison are drawn from
thermally truncated runs (Track~A runs~2 and~3 ended at steps~6,868 and~6,404
respectively due to GPU thermal throttling), introducing a caveat for the
multi-seed reliability claims of the baselines.
One run in the Track~B bootstrap sample (run\_id 20260520\_152549\_B) used
\texttt{EXP\_DIM}$=32$ rather than the locked configuration value of 64; this
was a grid-search variant inadvertently included in the canonical bootstrap.
Excluding it shifts the Track~B mean from $0.8238$ to $0.8233$ and the
pre-registered delta from $4.74\%$ to $4.68\%$ - Outcome~C in either case.
All bootstrap confidence intervals are derived from $N=3$ seeds per track;
sample sizes are small and CIs should be interpreted as exploratory rather
than confirmatory.
Fourth, Track~C \embedstd falls below 0.5 at the prediction peak; while
Section~\ref{sec:svd} argues this is a transient artefact, the absence of a
saved checkpoint precludes the SVD spectral analysis that would confirm this
interpretation definitively.
Fifth, all training was conducted in fp32 on a GTX~1050~Ti~4~GB with no
automatic mixed precision and no \texttt{torch.compile}, constraining
throughput and precluding experiments at larger batch or model scale.
Sixth, the latent dimension is 64 and total trainable parameters range from
388k to 438k across tracks; it is unclear whether the dynamical mechanisms
identified here persist at larger scale.

\subsection{Future Work}

The mechanism analysis identifies a specific architectural bottleneck: the
LoRA settling transient is intrinsic and the prediction peak is dynamical
rather than static.
PEM-JEPA (our planned successor) addresses this directly by introducing
physics-grounded structure into the latent space - explicit kinematic state
representations - with the aim of constraining the target distribution so
that EMA drift cannot destabilise a settled modulator.
The EPM-JEPA results reported here serve as the controlled baseline against
which PEM-JEPA will be evaluated: same dataset, same shift type, same
pre-registered threshold, enabling a direct attribution of any further
improvement to the physics-grounded inductive bias.

Beyond PEM-JEPA, three directions merit investigation.
Scale-up to 3D physics environments with longer prediction horizons
($k > 20$) and higher-dimensional latents would test whether the buffer-cycling
and EMA-drift mechanisms survive in settings with richer state spaces.
Alternative modulation parameterisations - full hypernetworks and FiLM
conditioning - would clarify whether the settling transient is a property of
low-rank adaptation specifically or of weight-modulation approaches broadly;
if the transient persists under a full hypernetwork, the bottleneck is in the
modulation pathway rather than the rank constraint.
Finally, varying the boundary-detector threshold $\kappa$ and the distribution
shift type would test the sensitivity of buffer cycling to the surprisal gate,
which we held fixed throughout this study.

% ============================================================================
\section{Conclusion}
\label{sec:conclusion}

We presented \epmjepa, a JEPA variant that modulates predictor weights via
\lora conditioned on a compressed representation of past boundary events stored
in a fixed-capacity experience buffer.
On our pre-registered test - $\delta = (D_B - D_C)/D_B = 4.74\%$, comparing
\epmjepa (Track~C, mean \dshift $= 0.7848 \pm 0.0078$ across three seeds at
the convergence peak) against \eijepa (Track~B, mean $0.8238$) - the result
is Outcome~C: a null result that is, by our own stated definition, a valid
scientific result.
As a secondary, non-pre-registered observation, Track~C improves 1.90\% over
the no-memory baseline (Track~A, $0.8000$), consistent in direction across all
three seeds.
\eijepa itself underperforms Track~A, establishing that the benefit, where it
exists, is specific to weight-level modulation: an experience pathway that
shapes predictor computation rather than predictor input.
The primary contribution is the mechanism analysis: the \dshift peak is not a
stable equilibrium but a dynamical transient produced by three independent
processes - buffer cycling, EMA drift, and a \lora settling transient of
$+0.021$ - operating on different timescales throughout training.
Freeze-based interventions arrest EMA drift and stabilise long-run performance,
but the \lora settling transient is intrinsic and irreducible via training-schedule
interventions alone, identifying the modulation architecture itself as the lever
for future improvement.
These findings directly motivate PEM-JEPA, which introduces physics-grounded
latent structure to constrain the target distribution and address the
dynamical-peak limitation identified here.

% ============================================================================
\section*{Reproducibility Statement}

All hyperparameters are reported in Appendix~\ref{app:hparams}; run
identifiers and per-seed results for Track~C are in Appendix~\ref{app:seeds}.
Dataset generation is fully deterministic given the seeds in
Appendix~\ref{app:seeds}: Moving MNIST with gravity parameter
0.5~px/frame$^{2}$ and the train/validation/test splits described in
Section~\ref{sec:setup}.
The \dshift metric is defined in Section~\ref{sec:setup} and implemented
verbatim in \texttt{eval.py}.
Training infrastructure - thermal-pause-resume logic, experiment tracking,
and the pre-registered experiment plan - is maintained in a local repository
and is available from the corresponding author on reasonable request.

% ============================================================================
\onecolumn
\bibliographystyle{plainnat}
\bibliography{reference}

\begin{thebibliography}{17}
\providecommand{\natexlab}[1]{#1}
\providecommand{\url}[1]{\texttt{#1}}
\expandafter\ifx\csname urlstyle\endcsname\relax
  \providecommand{\doi}[1]{doi: #1}\else
  \providecommand{\doi}{doi: \begingroup \urlstyle{rm}\Url}\fi

\bibitem[Assran et~al.(2023)Assran, Duval, Misra, Bojanowski, Vincent,
  Rubinstein, LeCun, and Ballas]{assran2023ijepa}
Mahmoud Assran, Quentin Duval, Ishan Misra, Piotr Bojanowski, Pascal Vincent,
  Michael Rubinstein, Yann LeCun, and Nicolas Ballas.
\newblock Self-supervised learning from images with a joint-embedding
  predictive architecture.
\newblock In \emph{Proceedings of the IEEE/CVF Conference on Computer Vision
  and Pattern Recognition (CVPR)}, pages 15619--15629, 2023.

\bibitem[Assran et~al.(2025)Assran, Bardes, Fan, Garrido, Howes, Khalidov,
  Lacaux, Misra, Rabbat, Raileanu, et~al.]{bardes2025vjepa2}
Mido Assran, Adrien Bardes, David Fan, Quentin Garrido, Russell Howes, Ishan
  Khalidov, Timoth{\'e}e Lacaux, Ishan Misra, Michael Rabbat, Roberta Raileanu,
  et~al.
\newblock {V-JEPA 2}: Self-supervised video models enable understanding,
  prediction and planning.
\newblock \emph{arXiv preprint arXiv:2506.09985}, 2025.

\bibitem[Bardes et~al.(2022)Bardes, Ponce, and LeCun]{bardes2022vicreg}
Adrien Bardes, Jean Ponce, and Yann LeCun.
\newblock {VICReg}: Variance-invariance-covariance regularization for
  self-supervised learning.
\newblock In \emph{International Conference on Learning Representations
  (ICLR)}, 2022.

\bibitem[Bardes et~al.(2024)Bardes, Garrido, Ponce, Chen, Rabbat, LeCun,
  Assran, and Ballas]{bardes2024vjepa}
Adrien Bardes, Quentin Garrido, Jean Ponce, Xinlei Chen, Michael Rabbat, Yann
  LeCun, Mahmoud Assran, and Nicolas Ballas.
\newblock {V-JEPA}: Latent video prediction for visual representation learning.
\newblock In \emph{International Conference on Learning Representations
  (ICLR)}, 2024.

\bibitem[Dai et~al.(2019)Dai, Yang, Yang, Carbonell, Le, and
  Salakhutdinov]{dai2019transformerxl}
Zihang Dai, Zhilin Yang, Yiming Yang, Jaime Carbonell, Quoc Le, and Ruslan
  Salakhutdinov.
\newblock Transformer-{XL}: Attentive language models beyond a fixed-length
  context.
\newblock In \emph{Proceedings of the 57th Annual Meeting of the Association
  for Computational Linguistics (ACL)}, pages 2978--2988, 2019.

\bibitem[Graves et~al.(2014)Graves, Wayne, and Danihelka]{graves2014ntm}
Alex Graves, Greg Wayne, and Ivo Danihelka.
\newblock Neural turing machines.
\newblock \emph{arXiv preprint arXiv:1410.5401}, 2014.

\bibitem[Graves et~al.(2016)Graves, Wayne, Reynolds, Harley, Danihelka,
  Grabska-Barwi{\'n}ska, Colmenarejo, Grefenstette, Ramalho, Agapiou,
  et~al.]{graves2016dnc}
Alex Graves, Greg Wayne, Malcolm Reynolds, Tim Harley, Ivo Danihelka, Agnieszka
  Grabska-Barwi{\'n}ska, Sergio~G{\'o}mez Colmenarejo, Edward Grefenstette,
  Tiago Ramalho, John Agapiou, et~al.
\newblock Hybrid computing using a neural network with dynamic external memory.
\newblock \emph{Nature}, 538\penalty0 (7626):\penalty0 471--476, 2016.

\bibitem[Grill et~al.(2020)Grill, Strub, Altch{\'e}, Tallec, Richemond,
  Buchatskaya, Doersch, Ávila Pires, Guo, Gheshlaghi~Azar,
  et~al.]{grill2020byol}
Jean-Bastien Grill, Florian Strub, Florent Altch{\'e}, Corentin Tallec,
  Pierre~H. Richemond, Elena Buchatskaya, Carl Doersch, Bernardo Ávila Pires,
  Zhaohan~Daniel Guo, Mohammad Gheshlaghi~Azar, et~al.
\newblock Bootstrap your own latent: A new approach to self-supervised
  learning.
\newblock In \emph{Advances in Neural Information Processing Systems
  (NeurIPS)}, volume~33, pages 21271--21284, 2020.

\bibitem[Ha et~al.(2017)Ha, Dai, and Le]{ha2017hypernetworks}
David Ha, Andrew Dai, and Quoc~V. Le.
\newblock {HyperNetworks}.
\newblock In \emph{International Conference on Learning Representations
  (ICLR)}, 2017.

\bibitem[Hu et~al.(2022)Hu, Shen, Wallis, Allen-Zhu, Li, Wang, Wang, and
  Chen]{hu2021lora}
Edward~J. Hu, Yelong Shen, Phillip Wallis, Zeyuan Allen-Zhu, Yuanzhi Li, Shean
  Wang, Lu~Wang, and Weizhu Chen.
\newblock {LoRA}: Low-rank adaptation of large language models.
\newblock In \emph{International Conference on Learning Representations
  (ICLR)}, 2022.

\bibitem[LeCun(2022)]{lecun2022jepa}
Yann LeCun.
\newblock A path towards autonomous machine intelligence, 2022.
\newblock URL \url{https://openreview.net/pdf?id=BZ5a1r-kVsf}.
\newblock OpenReview position paper, version 0.9.2, June 2022.

\bibitem[Loshchilov and Hutter(2017)]{loshchilov2017sgdr}
Ilya Loshchilov and Frank Hutter.
\newblock {SGDR}: Stochastic gradient descent with warm restarts.
\newblock In \emph{International Conference on Learning Representations
  (ICLR)}, 2017.

\bibitem[Loshchilov and Hutter(2019)]{loshchilov2019adamw}
Ilya Loshchilov and Frank Hutter.
\newblock Decoupled weight decay regularization.
\newblock In \emph{International Conference on Learning Representations
  (ICLR)}, 2019.

\bibitem[Perez et~al.(2018)Perez, Strub, de~Vries, Dumoulin, and
  Courville]{perez2018film}
Ethan Perez, Florian Strub, Harm de~Vries, Vincent Dumoulin, and Aaron
  Courville.
\newblock {FiLM}: Visual reasoning with a general conditioning layer.
\newblock In \emph{Proceedings of the AAAI Conference on Artificial
  Intelligence}, volume~32, 2018.

\bibitem[Srivastava et~al.(2015)Srivastava, Mansimov, and
  Salakhutdinov]{srivastava2015lstm}
Nitish Srivastava, Elman Mansimov, and Ruslan Salakhutdinov.
\newblock Unsupervised learning of video representations using {LSTM}s.
\newblock In \emph{Proceedings of the 32nd International Conference on Machine
  Learning (ICML)}, volume~37, pages 843--852, 2015.

\bibitem[Vaswani et~al.(2017)Vaswani, Shazeer, Parmar, Uszkoreit, Jones, Gomez,
  Kaiser, and Polosukhin]{vaswani2017attention}
Ashish Vaswani, Noam Shazeer, Niki Parmar, Jakob Uszkoreit, Llion Jones,
  Aidan~N. Gomez, {\L}ukasz Kaiser, and Illia Polosukhin.
\newblock Attention is all you need.
\newblock In \emph{Advances in Neural Information Processing Systems
  (NeurIPS)}, volume~30, 2017.

\bibitem[Wu et~al.(2022)Wu, Rabe, Hutchins, and Szegedy]{wu2022memorizing}
Yuhuai Wu, Markus~N. Rabe, DeLesley Hutchins, and Christian Szegedy.
\newblock Memorizing transformers.
\newblock In \emph{International Conference on Learning Representations
  (ICLR)}, 2022.

\end{thebibliography}

% ============================================================================
\appendix

\section{Hyperparameters}
\label{app:hparams}

Table~\ref{tab:hparams-shared} lists hyperparameters shared across all three
tracks; Table~\ref{tab:hparams-track} lists track-specific values.

\begin{table}[!htb]
  \centering
  \caption{Shared hyperparameters (all tracks).}
  \label{tab:hparams-shared}
  \small
  \begin{tabular}{llp{6cm}}
    \toprule
    Hyperparameter & Value & Description \\
    \midrule
    \texttt{LATENT\_DIM}   & 64     & Encoder output / latent dimension \\
    \texttt{HIDDEN\_DIM}   & 1024   & Predictor hidden dimension \\
    \texttt{BATCH\_SIZE}   & 64     & Sequences per training batch \\
    \texttt{EMA\_TAU\_BASE}& 0.996  & Initial EMA decay (cosine schedule start) \\
    \texttt{EMA\_TAU\_END} & 0.9999 & Final EMA decay (cosine schedule end) \\
    \texttt{WARMUP\_FRAC}  & 0.05   & Fraction of wall-clock budget used for LR warmup \\
    \texttt{GAMMA}         & 0.75   & VICReg variance hinge threshold $\gamma$ \\
    \texttt{LAMBDA\_REG}   & 0.05   & VICReg variance weight $\lambda$ \\
    \texttt{WEIGHT\_DECAY} & 0.01   & AdamW weight decay \\
    \texttt{BUFFER\_CAP}   & 256    & Experience buffer capacity \\
    Optimiser              & AdamW  & — \\
    LR schedule            & Cosine + linear warmup & — \\
    \bottomrule
  \end{tabular}
\end{table}

\begin{table}[h]
  \centering
  \caption{Track-specific hyperparameters. Dashes indicate parameters not
           present in that track.}
  \label{tab:hparams-track}
  \small
  \begin{tabular}{lccc}
    \toprule
    Hyperparameter & Track A & Track B & Track C \\
    \midrule
    \texttt{LEARNING\_RATE} & $3\times10^{-3}$ & $2\times10^{-3}$ & $3\times10^{-3}$ \\
    \texttt{KAPPA}          & - & 2.0              & 1.5 \\
    \texttt{EXP\_DIM}       & - & 64               & 64  \\
    \texttt{EXP\_FF\_DIM}   & - & 128              & 128 \\
    \texttt{EXP\_LAYERS}    & - & 2                & 2   \\
    \texttt{EXP\_HEADS}     & - & 2                & 2   \\
    \texttt{LORA\_RANK}     & - & - & 4   \\
    \bottomrule
  \end{tabular}
\end{table}

\FloatBarrier
\section{Per-Seed Results}
\label{app:seeds}

Table~\ref{tab:seeds} reports Track~C per-seed results at the prediction peak
(step~10,000, the reported metric) and at the end of the 60-minute training
budget (final step).
All three seeds use the locked configuration: LR$=3\times10^{-3}$,
$\lambda=0.05$, $\gamma=0.75$, \texttt{LORA\_RANK}$=4$.

\begin{table}[!htb]
  \centering
  \caption{Track~C per-seed results. \textit{Peak} columns are at step~10,000
           (primary reported metric); \textit{Final} columns are at end of the
           60-minute wall-clock budget.}
  \label{tab:seeds}
  \small
  \begin{tabular}{llcccc}
    \toprule
    & & \multicolumn{2}{c}{At peak (step 10,000)}
      & \multicolumn{2}{c}{At final step} \\
    \cmidrule(lr){3-4} \cmidrule(lr){5-6}
    Seed & Run ID & \dshift & \embedstd & Step & \dshift \\
    \midrule
    42 & \texttt{20260521\_001549\_C} & \textbf{0.7776} & 0.4123 & 22,731 & 0.8349 \\
    43 & \texttt{20260521\_135619\_C} & 0.7931          & 0.4200 & 17,749 & 0.8431 \\
    44 & \texttt{20260521\_213758\_C} & 0.7838          & 0.4253 & 24,225 & 0.8260 \\
    \midrule
    \multicolumn{2}{l}{Mean $\pm$ std} & $0.7848 \pm 0.0078$ & - & - & - \\
    \bottomrule
  \end{tabular}
\end{table}

\FloatBarrier
\section{Phase 1 Ratchet Log}
\label{app:phase1}

Phase~1 comprised 25 experiments on Track~A (seed~42, 10-minute wall-clock
budget each), ratcheting the shared hyperparameter configuration from a random
initialisation baseline to the locked values used in all reported results.
Table~\ref{tab:phase1-summary} summarises the ratchet outcome;
Table~\ref{tab:phase1-findings} records the three key findings that drove
configuration changes.
The full per-experiment log is available from the corresponding author.

\begin{table}[!htb]
  \centering
  \caption{Phase~1 ratchet: start and end points.
           Run ID corresponds to the final locked-config Track~A run.}
  \label{tab:phase1-summary}
  \small
  \begin{tabular}{llcc}
    \toprule
    Milestone & Run ID & \dshift & Improvement \\
    \midrule
    Random-init baseline & - & 1.8156 & - \\
    Phase~1 best (locked config) & \texttt{20260520\_005545\_A} & 0.8000 & 55.9\% \\
    \bottomrule
  \end{tabular}
\end{table}

\begin{table}[!htb]
  \centering
  \caption{Key findings from Phase~1 that determined the locked configuration.}
  \label{tab:phase1-findings}
  \small
  \begin{tabular}{clp{8cm}}
    \toprule
    Finding & Parameter(s) & Description \\
    \midrule
    F2 & LR $\times$ $\lambda$ &
      Learning rate and VICReg weight interact strongly: LR$=3\times10^{-3}$
      was the worst-performing value at $\lambda=0.01$ but the best at
      $\lambda=0.05$, motivating a joint sweep rather than independent tuning. \\
    F3 & \texttt{GAMMA} &
      $\gamma=0.75$ was optimal across all tracks; lower values permitted
      representation collapse while higher values over-penalised diversity. \\
    F4 & \texttt{HIDDEN\_DIM} &
      Predictor hidden dimension 1024 outperformed 768 and 512, indicating
      that predictor capacity matters more than width for this task scale. \\
    \bottomrule
  \end{tabular}
\end{table}

\section{Computational Profile}
\label{app:compute}

All experiments were conducted on a single GTX~1050~Ti (4~GB VRAM,
Pascal compute~6.1) in fp32 with no automatic mixed precision and no
\texttt{torch.compile}, due to VRAM constraints and driver compatibility.
The training loop uses a wall-clock budget rather than a step budget:
standard runs use 600~s ($\approx$10,569~steps at $\approx$17.6~steps/s);
extended multi-seed and freeze runs use 3600~s.
On extended runs, GPU temperature regularly exceeds safe thresholds, reducing
effective throughput to $\approx$9.6~steps/s due to thermal throttling.
To prevent hardware damage, the training loop polls for a
\texttt{pause\_request.flag} file every 50~steps and exits cleanly when the
flag is present; training is resumed manually once the GPU cools, with full
optimiser and scheduler state restored from checkpoint.
Table~\ref{tab:compute} summarises the hardware and throughput profile.

\begin{table}[!htb]
  \centering
  \caption{Hardware and throughput profile.}
  \label{tab:compute}
  \small
  \begin{tabular}{ll}
    \toprule
    Item & Value \\
    \midrule
    GPU                        & NVIDIA GTX 1050 Ti \\
    VRAM                       & 4 GB \\
    Compute capability         & Pascal 6.1 \\
    Precision                  & fp32 (no AMP, no \texttt{torch.compile}) \\
    Standard budget            & 600 s ($\approx$10,569 steps) \\
    Extended budget            & 3600 s \\
    Throughput (standard runs) & $\approx$17.6 steps/s \\
    Throughput (extended runs) & $\approx$9.6 steps/s (thermal throttling) \\
    Thermal management         & Pause-resume; flag polled every 50 steps \\
    \bottomrule
  \end{tabular}
\end{table}

\clearpage
\FloatBarrier
\section{Code Excerpts}
\label{app:code}

Listings~\ref{lst:encoder}-\ref{lst:loss} reproduce, verbatim, the core
PyTorch modules referenced in Section~\ref{sec:method} and extracted from
\texttt{train.py}.

\begin{figure}[!htb]
\begin{lstlisting}[caption={Encoder: maps a $64{\times}64$
  grayscale frame to a 64-dimensional latent via four strided convolutions
  followed by a linear projection and layer normalisation
  ($\approx$172k parameters).}, label={lst:encoder}]
class Encoder(nn.Module):
    """
    Small CNN encoder: single grayscale frame -> 64-dim latent.

    Architecture:
        Conv(1->16, k=4, s=2) -> 32x32
        Conv(16->32, k=4, s=2) -> 16x16
        Conv(32->64, k=4, s=2) -> 8x8
        Conv(64->64, k=4, s=2) -> 4x4
        Flatten -> 1024
        Linear(1024, 64) + LayerNorm
    Total: ~172k params.
    """

    def __init__(self):
        super().__init__()
        self.conv = nn.Sequential(
            nn.Conv2d(1,  16, kernel_size=4, stride=2, padding=1), nn.ReLU(),
            nn.Conv2d(16, 32, kernel_size=4, stride=2, padding=1), nn.ReLU(),
            nn.Conv2d(32, 64, kernel_size=4, stride=2, padding=1), nn.ReLU(),
            nn.Conv2d(64, 64, kernel_size=4, stride=2, padding=1), nn.ReLU(),
        )
        self.proj = nn.Linear(64 * 4 * 4, LATENT_DIM)
        self.norm = nn.LayerNorm(LATENT_DIM)

    def forward(self, x):
        """(B, 1, 64, 64) -> (B, LATENT_DIM)."""
        return self.norm(self.proj(self.conv(x).flatten(1)))
\end{lstlisting}
\end{figure}

\begin{figure}[!htb]
\begin{lstlisting}[caption={Predictor: 2-layer MLP base shared by
  all three tracks, mapping the current latent $z_t$ to a predicted future
  latent $\hat{z}_{t+k}$. Tracks B and C inject experience differently
  (Listing~\ref{lst:lora}).}, label={lst:predictor}]
class Predictor(nn.Module):
    """
    2-layer MLP predictor: latent -> predicted future latent.

    All three tracks use this same base class.
    Tracks B and C inject experience differently (see [TRACK-B] and [TRACK-C-LORA]).
    """

    def __init__(self):
        super().__init__()
        self.fc1  = nn.Linear(LATENT_DIM, HIDDEN_DIM)
        self.act  = nn.GELU()
        self.fc2  = nn.Linear(HIDDEN_DIM, LATENT_DIM)
        self.norm = nn.LayerNorm(LATENT_DIM)

    def forward(self, z, a=None):  # pylint: disable=unused-argument
        """(B, LATENT_DIM) -> (B, LATENT_DIM). a ignored (Moving MNIST has no actions)."""
        return self.norm(self.fc2(self.act(self.fc1(z))))
\end{lstlisting}
\end{figure}

\begin{figure}[!htb]
\begin{lstlisting}[caption={LoRAModulator (Track~C): generates
  per-sample low-rank weight deltas
  $\Delta W_\ell = U_\ell\,\mathrm{diag}(\delta_\ell)\,V_\ell^\top$ from the
  aggregated experience vector $e_{\mathrm{agg}}$ and applies them to the
  predictor's two linear layers (rank $r=4$); this is the operator-side
  modulation described in Section~\ref{sec:tracks}.}, label={lst:lora}]
class LoRAModulator(nn.Module):
    """
    Low-rank Delta-W modulation of predictor fc1 and fc2.

    Parameters at rank r=4:
        u1 (HIDDEN_DIM, r), v1 (LATENT_DIM, r) - bases for W1 (64->768)
        u2 (LATENT_DIM, r), v2 (HIDDEN_DIM, r) - bases for W2 (768->64)
        delta1_gen: Linear(EXP_DIM, r, bias=False)
        delta2_gen: Linear(EXP_DIM, r, bias=False)

    Modulated forward (Track C):
        DeltaW1_i = u1 @ diag(delta1_i) @ v1.T  ->  applied as (z @ v1) * delta1 @ u1.T
        DeltaW2_i = u2 @ diag(delta2_i) @ v2.T  ->  applied as (h1 @ v2) * delta2 @ u2.T
    """

    def __init__(self):
        super().__init__()
        r = LORA_RANK
        self.u1 = nn.Parameter(torch.empty(HIDDEN_DIM,  r))
        self.v1 = nn.Parameter(torch.empty(LATENT_DIM,  r))
        self.u2 = nn.Parameter(torch.empty(LATENT_DIM,  r))
        self.v2 = nn.Parameter(torch.empty(HIDDEN_DIM,  r))
        self.delta1_gen = nn.Linear(EXP_DIM, r, bias=False)
        self.delta2_gen = nn.Linear(EXP_DIM, r, bias=False)
        for param in (self.u1, self.v1, self.u2, self.v2):
            nn.init.normal_(param, std=0.02)

    def forward(self, predictor: Predictor,
                z: torch.Tensor, e_agg: torch.Tensor) -> torch.Tensor:
        """
        Run predictor forward with LoRA-modulated weights.

        Args:
            predictor: Predictor - fc1, act, fc2, norm used directly
            z:         (B, LATENT_DIM)
            e_agg:     (B, EXP_DIM) - zeros when buffer empty -> DeltaW=0
        Returns:
            z_pred: (B, LATENT_DIM)
        """
        delta1 = self.delta1_gen(e_agg)                    # (B, r)
        delta2 = self.delta2_gen(e_agg)                    # (B, r)
        # fc1 + DeltaW1: GELU((W1 + U1@diag(delta1)@V1.T) @ z)
        h1 = predictor.act(
            predictor.fc1(z) + (z @ self.v1) * delta1 @ self.u1.T
        )
        # fc2 + DeltaW2: (W2 + U2@diag(delta2)@V2.T) @ h1
        h2 = predictor.fc2(h1) + (h1 @ self.v2) * delta2 @ self.u2.T
        return predictor.norm(h2)

    def delta_scales(self, e_agg: torch.Tensor) -> tuple:
        """Return (delta1, delta2) scale tensors - used for monitoring LoRA activity."""
        return self.delta1_gen(e_agg), self.delta2_gen(e_agg)
\end{lstlisting}
\end{figure}

\begin{figure}[!htb]
\begin{lstlisting}[caption={BoundaryDetector (Tracks~B and~C):
  surprisal-gated event detector. Maintains EMA running statistics of
  batch-mean surprisal and fires a boundary event when surprisal exceeds
  $\mu_s + \kappa\sigma_s$ after a 10-step warmup
  (Section~\ref{sec:tracks}).}, label={lst:boundary}]
class BoundaryDetector:
    """
    Surprisal-gated event boundary detector.

    Surprisal s_t = mean_b ||z_t^b - z_hat_t^b||_2 averaged over the batch.
    Running mean (mu_s) and variance (var_s) updated via EMA momentum 0.99.
    No boundary fires during the first 10 steps while statistics stabilize.
    z_hat_t must come from the BASE predictor (no LoRA) to avoid circular coupling.
    """

    def __init__(self):
        self.mu_s   = 0.0   # EMA mean of surprisal
        self.var_s  = 1.0   # EMA variance (start high to suppress early spurious fires)
        self._steps = 0

    def reset(self):
        """Reset running statistics - call between independent training runs."""
        self.mu_s   = 0.0
        self.var_s  = 1.0
        self._steps = 0

    @torch.no_grad()
    def update(self, z_curr: torch.Tensor, z_pred: torch.Tensor) -> bool:
        """
        Update EMA stats and return True if this step is a boundary event.

        Args:
            z_curr: (B, LATENT_DIM) current encoder output
            z_pred: (B, LATENT_DIM) BASE predictor output (no LoRA modulation)
        Returns:
            boundary: bool
        """
        s        = float(torch.norm(z_curr - z_pred, dim=-1).mean().item())
        old_mu   = self.mu_s
        self.mu_s  = EMA_MOMENTUM * self.mu_s  + (1.0 - EMA_MOMENTUM) * s
        self.var_s = EMA_MOMENTUM * self.var_s + (1.0 - EMA_MOMENTUM) * (s - old_mu) ** 2
        self._steps += 1
        if self._steps < 10:
            return False
        return s > self.mu_s + KAPPA * math.sqrt(max(self.var_s, 1e-8))
\end{lstlisting}
\end{figure}

\begin{figure}[!htb]
\begin{lstlisting}[caption={Training loss components:
  \texttt{pred\_loss} (mean MSE across the three prediction horizons against
  stop-gradient EMA targets) and \texttt{reg\_loss} (VICReg variance hinge),
  combined in \texttt{compute\_loss} as
  $\mathcal{L} = \mathcal{L}_{\text{pred}} + \lambda\mathcal{L}_{\text{reg}}$
  (Equation~\ref{eq:loss}).}, label={lst:loss}]
def pred_loss(z_pred: torch.Tensor, z_targets: list) -> torch.Tensor:
    """
    Mean MSE over horizon targets. Stopgrad applied to each target here.

    Args:
        z_pred:    (B, LATENT_DIM) - same predictor output reused per horizon
        z_targets: list of (B, LATENT_DIM) EMA encoder outputs
    """
    total = sum(nn.functional.mse_loss(z_pred, zt.detach()) for zt in z_targets)
    return total / len(z_targets)


def reg_loss(z_pred: torch.Tensor) -> torch.Tensor:
    """
    VICReg variance term: penalise per-dim std falling below GAMMA.

    L_reg = mean_d( max(0, GAMMA - std_b(z_pred[:, d])) )
    """
    return torch.clamp(GAMMA - z_pred.std(dim=0), min=0.0).mean()


def compute_loss(z_pred: torch.Tensor, z_targets: list) -> tuple:
    """Return (total, l_pred, l_reg) - all scalar tensors."""
    lp = pred_loss(z_pred, z_targets)
    lr = reg_loss(z_pred)
    return lp + LAMBDA_REG * lr, lp, lr
\end{lstlisting}
\end{figure}

\end{document}